\documentclass[runningheads]{llncs}
\usepackage{graphicx}

\usepackage{hyperref}

\begin{document}
\title{A Survey and Approach to Chart Classification}
%
%

\author{Anurag Dhote\inst{1}\orcidID{0009-0000-9385-4758} \and
Mohammed Javed\inst{1}\thanks{Corresponding author}\orcidID{0000-0002-3019-7401} \and
David S Doermann\inst{2}\orcidID{0000-0003-1639-4561}}

\institute{Indian Institute of Information Technology, Allahabad \and
Department of CSE, University at Buffalo, Buffalo, NY, USA
\email{Email:\{mit2021082@iiita.ac.in, javed@iiita.ac.in, 
   doermann@buffalo.edu\} 
  }\\
}

\authorrunning{A. Dhote et al.}
%
%
\maketitle              
\begin{abstract}
Charts represent an essential source of visual information in documents and facilitate a deep understanding and interpretation of information typically conveyed numerically. In the scientific literature, there are many charts, each with its stylistic differences. Recently the document understanding community has begun to address the problem of automatic chart understanding, which begins with chart classification. In this paper, we present a survey of the current state-of-the-art techniques for chart classification and discuss the available datasets and their supported chart types. We broadly classify these contributions as traditional approaches based on ML, CNN, and Transformers.

Furthermore, we carry out an extensive comparative performance analysis of CNN-based and transformer-based approaches on the recently published CHARTINFO UB-UNITECH PMC dataset for the CHART-Infographics competition at ICPR 2022. The data set includes 15 different chart categories, including 22,923 training images and 13,260 test images. We have implemented a vision-based transformer model that produces state-of-the-art results in chart classification. 

\end{abstract}
\section{Introduction}
Charts provide a compact summary of important information or research findings in technical documents and are a powerful visualization tool widely used by the scientific and business communities. In the recent literature, the problem of chart mining has attracted increased attention due to numerous advantages, as suggested in the comprehensive survey published by Davila et al. in 2019 \cite{davila_19_survey}. The term Chart mining refers to the process of extracting information represented by charts. Another motivating factor in the increased attention paid to this problem is a series of competitions held in conjunction with significant conferences to address the critical challenges in the chart mining pipeline\cite{davila_19,davila_20,davila_22}. 


Since a variety of charts are possible, chart classification is often the first step in chart mining. The task of chart image classification can be formalized as, given a chart image extracted from a document, classifying the image into one of $N$ defined categories. The wide variety of chart types in the literature adds to the complexity of the task\cite{chagas_18,davila_19_survey,thiyam_21}. Some additional problems include interclass similarity, noise in authentic chart images, and more state-of-the-art datasets that cover multiple chart types and incorporate 2.5 or 3D charts and noise into the training samples\cite{thiyam_21}. The rise of robust deep learning models has contributed significantly to the success of chart classification. Deep learning approaches have outperformed traditional machine learning approaches regarding robustness and performance. Yet there need to be more state-of-the-art solutions that can provide stable results and are robust enough to address noise in some data sets. In this paper, we provide a performance comparison of several deep learning models that are state-of-the-art in the ImageNet\cite{russakovsky_15} classification task. 
In addition, we report the performances of several popular vision transformers, which, to the best of our knowledge, have yet to be used for chart classification, except for the recent ICPR 2022 CHART-Infographics competition\cite{davila_22}.

This paper is organized as follows. Section 2 summarizes the existing chart classification literature covering traditional and deep learning-based methods, including a brief discussion on transformer-based chart classification. Section 3 reports and summarizes publicly available datasets.
Section 4 briefly highlights the popular ImageNet pre-trained deep learning-based models that will be used for our comparative study. Section 5 describes the latest edition of the UB PMC dataset, the training and testing protocols, and a discussion on their performance for chart classification. Section 6 provides information on possible improvements and suggestions for future research. Finally, Section 7 concludes with a summary of the paper.

\section{Chart Classification Techniques}

Based on the type of approaches used to implement the chart classification task in the literature, they can be grouped into traditional ML, CNN-based deep learning, and Transformer-based deep learning. Each type of approach is described briefly below.

 \subsection{Traditional ML approaches}

Traditional approaches rely on feature extraction methods that are often manual and general-purpose. Features are extracted and then represented in mathematical form for direct processing by machine learning classifiers. Savva et al.\cite{savva_11} present a system that automatically reformats visualizations to increase visual comprehension. The authors use low-level image features for classification in conjunction with text-level features. The system uses a multiclass SVM classifier trained on a corpus containing 2601 chart images labeled with ten categories, following Gao et al.'s manual extraction approach. In \cite{gao_12}, researchers propose VIEW, a system that automatically extracts information from raster-format charts. The authors used an SVM to separate the textual and graphical components and classify the chart images based on the graphic elements extracted from the visual components. The text is typically found in three chart categories - bar charts, pie charts, and line graphs, with 100 images for each category collected from various real-world digital resources. 

Instead of taking an image as input, Karthikeyani and Nagarajan\cite{karthikeyani_12} present a system to recognize chart images from PDF documents using eleven texture features that are part of a Gray Level Co-Occurrence Matrix. A chart image is located in the PDF Document database, and the features are extracted and fed to the learning model. SVM, KNN, and MLP are the classifiers used for classification. Cheng et al.\cite{cheng_13} employ a multimodal approach that uses text and image features. These features are provided as input to an MLP. The output is characterized as a fuzzy set to get the final result. The corpus contains 1707 charts with three categories and a 96.1\%\ classification result. 

\subsection{CNN-based Deep Learning Approaches}

Liu et al.\cite{liu_15} used a combination of Convolutional Neural Networks (CNNs) and Deep Belief networks (DBNs) to capture high-level information present in deep hidden layers. Fully Connected Layers of Deep CNN are used to extract deeply hidden features. A DBN is then used to predict the image class using the deep hidden features. The authors use transfer learning and perform fine-tuning to prevent overfitting. They use a data set that includes more than $5,000$ images of charts, including pie, scatter, line, bar, and flow classes. Deep features are useful over primitive features to provide better stability and scalability to the proposed framework. The proposed method achieves an average accuracy of 75.4\%, which is 2.8\% more than the method that uses only deep ConvNets. 

Given the results of CNN in the classification of natural images, Siegel et al.\cite{siegel_16} used two CNN-based architectures for chart classification. They evaluated AlexNet and ResNet-50, which are pre-trained on the ImageNet data set and then fine-tuned for chart classification. This transfer learning approach is prevalent in subsequent works addressing this particular problem. The proposed frameworks outperformed the state-of-the-art model at the time, such as ReVision, by a significant margin. ResNet-50 achieved the best classification accuracy of 86\% on a data set that contained more than 60000 images spread over seven categories. 

Amara et al.\cite{amara_17} proposed a CNN-based on LeNet to classify images from their corpus of 3377 images into 11 categories. The model comprises eight layers, one input layer, five hidden layers, one fully connected layer, and one output layer. The fully connected layer is used as a classifier, while the hidden layers are convolution and pooling layers designed to extract features automatically. A fully connected layer employs softmax activation to classify images into defined classes. For evaluation of the model's performance, an 80-20 split is performed on the data set for training and assessment. The proposed model performs better than the LeNet and pretrained LeNet architectures with an accuracy of 89.5\%. 

Jung et al. \cite{jung_17} present a classification method using the deep learning framework Caffe and evaluate its efficacy by comparing it with ReVision\cite{savva_11}. The authors use GoogLeNet\cite{szegedy_14_googlenet} for classification and compare its results with shallower networks like LeNet-1 and AlexNet\cite{krizhevsky_12_alexnet}. GoogLeNet outperforms LeNet-1 and AlexNet with an accuracy of 91.3\%. Five-fold cross-validation is used for calculating the accuracy on an image corpus with 737 - 901 images for each chart type. The test concludes that ChartSense provides higher classification accuracy for all chart types than ReVision.

With studies adapting the deep learning approach for chart image classification, a comparative study of traditional vs. CNN architectures was required. Chagas et al.\cite{chagas_18} provide a comparative analysis of conventional vs. CNN techniques. Authors evaluated CNN architectures (VGG19\cite{simonyan_15_vgg}, Resnet-50\cite{he_15_resnet}, and Inception-V3\cite{szegedy_15_v3}) for chart image classification for ten classes of charts. The performance is compared with conventional machine learning classifiers, Naive Bayes, HOG features combined with KNN, Support Vector Machines, and Random Forests. Pre-trained CNN models with fine-tuned last convolutional layers were used. The authors concluded that CNN models surpass traditional methods with an accuracy of 77.76\% (Resnet-50) and 76.77\% (Inception-V3) compared to 45.03\% (HOG + SVM).

 Dia et al.\cite{dai_18} employ four deep learning models on a corpus of 11,174 chart images of five categories. Of AlexNet\cite{krizhevsky_12_alexnet}, VGG16\cite{simonyan_15_vgg}, GoogLeNet\cite{szegedy_14_googlenet} and ResNet\cite{he_15_resnet}, the authors get the best accuracy of 99.55\% for VGG16 model. VGG16 outperforms the models used in ChartSense paper by a large margin.

Significant roadblocks to chart mining research are caused by the fact that current chart data sets must be larger and contain sufficient diversity to support deep learning. To address this problem, Jobin et al.\cite{kv_19} presented DocFigure, a chart classification data set with $33,000$ charts in 28 different classes. To classify charts, the author's proposed techniques utilize deep features, deep texture features, and a combination of both. Among these baseline classification techniques, the authors observed that combining deep features and deep texture features classifies images more efficiently than individual features. The average classification accuracy improved by 3.94\% and 2.10\% by concatenating FC-CNN and FV-CNN over individual use of FC-CNN and FV-CNN, respectively. The overall accuracy of the combined feature methods turned out to be 92.90\%.

Luo et al. proposed a unified method to handle various chart styles\cite{luo_21}, where they show that generalization can be obtained in deep learning frameworks with rule-based methods. The experiments were performed on three different datasets of over $300,000$ images with three chart categories. In addition to the framework, an evaluation metric for the bar, line, and pie charts is also introduced. The authors concluded that the proposed framework performs better than traditional rules-based and pure deep learning methods. 

Araújo et al.\cite{araujo_20} implemented four classic CNN models that performed well on computer vision tasks, including Xception\cite{chollet_17_xception}, VGG19\cite{simonyan_15_vgg}, ResNet152\cite{he_15_resnet} and MobileNet\cite{howard_17_mobilenet}. The weights of these models were pre-trained on the ImageNet dataset, and the authors further performed hyperparameter tuning to obtain a stable learning rate and weight decay. These models were employed on a self-aggregated chart image corpus of 21,099 images with 13 different chart categories. Xception outperforms the other models by hitting an accuracy of 95\%. 

The problem of small datasets has been prevalent since the problem of chart mining was first introduced. Most work tries to increase the size of the dataset. However, Bajic and Job\cite{bajic_21} use a Siamese CNN network to work with smaller datasets. The authors show that an accuracy of 100\% can be achieved with 50 images per class, which is significantly better than using a vanilla CNN.     

With the increase in datasets for chart images and the rise of deep learning models being employed on said datasets, an empirical study of these deep learning models was due. Thiyam et al.\cite{thiyam_22} compared 15 different deep-learning models on a self-aggregated dataset of 110,182 images spfeatures24 different chart categories. In addition, the authors tested the performance of these models on several preexisting test sets. They concluded that Xception(90.25\%) and DenseNet121(90.12\%) provide the most consistent and stable performance of all the deep learning models. The authors arrived at this decision by employing a five-fold cross-validation technique and calculating the standard deviation for each model across all datasets.

 Davila et al.\cite{davila_19} summarized the work of different participants in the competition's first edition by harvesting raw tables from Infographics that provided data and tools for the chart recognition community. Two data sets were provided for the classification task. One was a synthetically generated AdobeSynth dataset, and the other UB-PMC data set was gathered from the PubMedCentral open-access library. The highest average F1-measure achieved for the synthetic data set was 99.81\% and the highest F1-measure achieved for the PMC data set was 88.29\%. In the second edition of the competition, the PMC set was improved and included in the training phase. An ensemble of ResNet152 and DenseNet121 achieved the highest F1-score of 92.8\%. The third edition of the competition was recently held at ICPR 2022. The corpus of real chart images was made up of 36,183 chart images. The winning team achieved an F1 score of 91\% with a base Swin transformer model with a progressive resizing technique. We summarize the competition details in Table \ref{tab:UBPMC}   
 
\begin{table*}[h]
    \centering
    \caption{Competition on Harvesting Raw Tables from Infographics (CHART-Infographics)}
    \label{tab:UBPMC}
    \resizebox{\linewidth}{!}{
    \begin{tabular}{|l|l|c|c|c|l|c|}
    \hline
     \textbf{Competition} & \textbf{Dataset} & \textbf{\#Classes} & \textbf{Train} & \textbf{Test} & \textbf{Top performing} & \textbf{F1-measure}\\
     & & & \textbf{Size} & \textbf{Size} & \textbf{Model} & \\
     \hline
     \hline
      ICDAR 2019 \cite{davila_19} & AdobeSynth & 10 & 198,010 & 4540 & ResNet-101 & 99.81\% \\ 
        & PMC & 7 & & 4242 & & 88.29\% \\
     \hline
      ICPR 2020 \cite{davila_20} & Adobe Synth & 12 & 14,400 & 2,999 & DenseNet-121 + & 100\% \\
       & UB PMC & 15 & 15,636 & 7,287 & ResNet-152 & 92.8\% \\
     \hline
     ICPR 2022  \cite{davila_22} & UB PMC & 15 & 22,923 & 13,620 & Swin Transformer & 91\% \\
     \hline
    \end{tabular}}
\end{table*}

\subsection{Transformer-based Deep Learning Approaches}
Since the inception of Vision Transformer, there has been a lot of development in various computer vision tasks such as image classification, object detection, and image segmentation. Vision transformer has outperformed CNN-based models in these tasks on the ImageNet dataset. However, there has not been widespread application of vision transformers to chart image classification. 
To our knowledge, only the Swin transformer\cite{liu_21_swin} has been used for chart classification as reported in \cite{davila_22}, which won the CHART-Infographics challenge ICPR2022. The authors applied a Swin Transformer Base Model with a progressive resizing technique. The models were initially trained on a scale (input size) of 224 followed by 384\cite{davila_22}.

The existing models in the literature are summarised in Table 2.

\begin{table*} [t]
\centering
\label{tab:models}
\caption{Published Literature on Chart Classification}
\resizebox{\linewidth}{!}{%
\begin{tabular}{|l|l|l|l|c|}
\hline
\textbf{Authors} & \textbf{Dataset} & \textbf{Model} & \textbf{Metric} & \textbf{Performance} \\
\hline
\hline
Savva et al.\cite{savva_11} & Self-acquired & SVM & Accuracy & 96.00\%\ \\ 
\hline
Gao et al.\cite{gao_12} & Self-acquired & SVM & Accuracy & 97.00\%\ \\
\hline
Kartikeyani &  & MLP & Accuracy & 69.68\%\ \\ 
and & Self-acquired & KNN &  & 78.06\%\ \\
Nagarajan\cite{karthikeyani_12} & & SVM &  & 76.77\%\ \\ 
\hline
Cheng et al.\cite{cheng_13}& Self-acquired & MLP & Accuracy & 96.10\% \\
\hline
Liu et al.\cite{liu_15} & DeepChart & CNN + DBN & Accuracy & 75.40\%\ \\
\hline
Siegel et al.\cite{siegel_16} & ChartSeer & AlexNet & Accuracy & 84.00\%\ \\ 
& & ResNet-50 & & 86.00\%\ \\
\hline
Amara et al.\cite{amara_17} & Self-acquired & CNN & Accuracy & 89.50\% \\
\hline
Jung et al.\cite{jung_17} &
Chart-Sense & GoogleNet & Accuracy & 91.30\%\ \\
\hline
Balaji et al.\cite{balaji_18} & Self-acquired & CNN & Accuracy & 99.72\%\ \\
\hline
Chagas et al.\cite{chagas_18} & Chart-Vega & ResNet-50 & Accuracy & 76.76\%\ \\ 
 & & Inception-V3 &  & 76.77\%\ \\
\hline
Dai et al.\cite{dai_18} & Self-acquired & ResNet & Accuracy & 98.89\%\ \\ 
& & GoogLeNet &  & 99.07\%\ \\
& & AlexNet &  & 99.48\%\ \\
& & VGG-16 &  & 99.55\%\ \\
\hline
Liu et al.\cite{liu_19} & Self-acquired & VGG-16 & Accuracy & 96.35\%\ \\ 
\hline
Davila et al.\cite{davila_19} & Synthetic & ResNet-101 & F1-measure & 99.81\%\ \\ 
& UB-PMC & ResNet-101 & & 88.29\% \\
\hline
Jobin et al.\cite{kv_19} & DocFigure & FC-CNN + FV-CNN & Accuracy & 91.30\%\ \\ 
\hline
Bajic et al.\cite{bajic_20} & Self-acquired & VGG-16 & Accuracy & 89.00\%\ \\ 
\hline
Araujo et al.\cite{araujo_20} & Self-acquired & Xception & Accuracy & 95.00\%\ \\
\hline
Luo et al.\cite{luo_21} & Chart-OCR & CNN & Custom(Bar) & 91.90\% \\
& & & Custom(Pie) & 91.80\% \\
& & & Custom(Line) & 96.20\% \\
\hline
Davila et al.\cite{davila_20} & UB-PMC & DenseNet-121 + ResNet-152 & F1-measure & 92.80\%\ \\ 
\hline 
Bajic and Job\cite{bajic_21} & Self-acquired & Siamese CNN & Accuracy & 100\%\ \\ 
\hline
Thiyam et al.\cite{thiyam_22} & Self-acquired & Xception & Accuracy & 90.25\%\ \\ 
& & DenseNet121 & & 90.12\%  \\
& & DenseNet201 &  & 90.53\% \\
\hline
Davila et al.\cite{davila_22} & UB-PMC & Swin Transformer & F1-measure & 91.00\%\ \\ 
\hline
\end{tabular}}
\end{table*}

\section{Chart Classification Datasets}
There has been a significant increase in the size of datasets both in terms of the number of samples and the number of chart types. The Revision dataset\cite{savva_11}  had only 2,601
images and 10 chart types. The recent publicly available dataset\cite{davila_22} comprises around 33,000 chart images of 15 different categories. The details of several publicly available datasets are discussed in this section.

\begin{table*} [ht]
\centering
\label{tab:datasets}
\caption{Chart Classification Datasets}
\resizebox{\linewidth}{!}{%
\begin{tabular}{|l|c|l|c|c|}
\hline
 \textbf{Dataset} & \textbf{Year} & \textbf{\#Samples} & \textbf{\#Category} & \textbf{Public} \\
  &  &  &  & \textbf{(Y/N)} \\
\hline
\hline
ReVision\cite{savva_11} & 2011 & 2601 & 10 & Y \\ 
\hline
View\cite{gao_12} & 2012 & 300 & 3 & N \\
\hline
Self\cite{karthikeyani_12} & 2012 & 155 & 8 & N\\
\hline
Self\cite{cheng_13} & 2014 & 1707 & 3 & N\\
\hline
DeepChart\cite{liu_15} & 2015 & 5000 & 5 & Y \\
\hline
ChartSeer\cite{siegel_16} & 2016 & 60000 & 7 & N \\ 
\hline
Self\cite{amara_17} & 2017 & 3377 & 11 & N \\
\hline
Chart-Sense\cite{jung_17} & 2017 & 6997 & 10 & Y  \\
\hline
Chart-Text\cite{balaji_18} & 2018 & 6000 & 2 & N \\
\hline
Chart-Vega\cite{chagas_18} & 2018 & 14471 & 10 & Y\\ 
\hline

Chart decoder\cite{dai_18} & 2018 & 11,174 & 5 & N \\
\hline
Self\cite{liu_19} & 2019 & 2500 & 2 & N \\ 
\hline
Synthetic\cite{davila_19} & 2019 & 202550 & 10 & Y \\
UB-PMC \cite{davila_19} & & 4242 & 7 & Y\\
\hline
DocFigure\cite{kv_19} & 2019 & 33000 & 28 & Y \\ 
\hline
Self\cite{bajic_20} & 2020 & 2702 & 10 & N \\ 
\hline
Self\cite{araujo_20} & 2020 & 21099 & 13 & N \\
\hline
Chart-OCR\cite{luo_21} & 2021  & 386966 & 3 & N \\ 
\hline
UB-PMC\cite{davila_20} & 2021  & 22924 & 15 & Y \\ 
\hline 
Self\cite{bajic_21} & 2021  & 3002 & 10 & N \\ 
\hline
Self\cite{thiyam_22} & 2021 & 110182 & 24 & N \\ 
\hline
UB-PMC\cite{davila_22} & 2022 & 33186 & 15 & Y\\ 
\hline

\end{tabular}}
\end{table*}

\paragraph{\textbf{ChartSense \cite{jung_17}:}} 
The ChartSense dataset was put together using the ReVision dataset, and the authors manually added some additional charts. The corpus has 5659 chart images that cover ten chart categories. 
\paragraph{\textbf{ChartVega \cite{chagas_18}:}}
This dataset has  ten chart types and was created due to a need for a benchmark dataset for chart image classification\cite{chagas_18}. The dataset contains both synthetic and real chart images. The set contains 14,471 chart images, of which 12059 are for training and 2412 are for testing. In addition, a validation set of 2683 real chart images is provided. No separate annotations are provided, as chart images are separated according to their types.

\paragraph{\textbf{DocFigure  \cite{kv_19}: }}
This corpus consists of 28 categories of annotated figure images. There are 33,000 images that include non-chart categories like natural images, tables, 3D objects, and medical images. The train set consists of 19,797 images, and the test set contains 13173 images. The labels are provided in a text document. 

\paragraph{\textbf{ChartOCR \cite{luo_21}:}}  The dataset contains 386,966 chart images created by the authors by crawling public excel sheets online. The dataset contains only three classes of chart images. The dataset is divided into the train, validation, and test sets. The training corpus contains 363,078 images, the validation set contains 11,932 images, and the test set contains 11,965 images. The annotations for the chart images are provided in JSON format.

\begin{table*} [t]
\centering
\caption{Composition of publicly available datasets}
\label{tab:Table3}
\resizebox{\linewidth}{!}{%
\begin{tabular}{|l |c| c |c |c| c|}
\hline
\textbf{Chart Type} & \textbf{UB-PMC } & \textbf{DocFigure } & \textbf{Chart-Sense } & \textbf{Chart-OCR} & \textbf{Chart-Vega } \\ 
  & \textbf{\cite{davila_22}} & \textbf{\cite{kv_19}} & \textbf{\cite{jung_17}} & \textbf{\cite{luo_21}} & \textbf{\cite{chagas_18}} \\ 
 \hline
 \hline
Arc & - & - & - & - & $1440$  \\ 

Area &  $308$ & $318$ & $509$ & - & $1440$  \\ 

Block & - & $1024$ & - & - & - \\ 

Bubble & - & $339$ & - & - & -\\ 

Flowchart & - & $1074$ & - & - & -\\ 

Heatmap & $377$ & $1073$ & - & - & -\\ 

Horizontal Bar &   $1421$ & - & - & - & -\\ 

Horizontal Interval &   $586$ & - & - & - & - \\ 

Line & $13956$ & $9022$ & $619$ & $122890$ & $1440$ \\ 

Manhattan &   $256$ & - & - & - & - \\ 

Map &   $906$ & $1078$ & $567$ & - & -\\ 

Parallel Coordinate & - & - & - & - & $1339$ \\ 

Pareto & - & $311$ & $391$ & - & -\\ 

Pie &   $433$ & $440$ & $568$ & $76922$ & $1440$\\ 

Polar & - & $338$ & - & - & -\\ 

Radar & - & $309$ & $465$ & - & -\\ 

Re-orderable Matrix & - & - & - & - & $1440$\\ 

Scatter & $2597$ & $1138$ & $696$ & & $1640$\\ 

Scatter-Line &   $3446$ & - & - & - & - \\ 

Sunburst & - & - & - & - & $1440$\\ 

Surface & $283$ & $395$ & - & - & -\\ 

Table & - & $1899$ & $594$ & - & -\\ 

Treemap & - & - & - & - & $1440$\\ 

Venn &   $206$ & $889$ & $693$ & - & -\\ 

Vertical Bar & $9199$ & $1196$ & $557$ & $187154$ & $1512$\\ 

Vertical Box &   $1538$ & $605$ & - & - & -\\ 

Vertical Interval &   $671$ & - & - & - & -\\ 
\hline

Total & $36183$ & $33071$ & $5659$ & $386966$ & $14471$ \\ \hline

\end{tabular}}

\end{table*}

\paragraph{\textbf{UB-PMC CHART-Infographics:}} This dataset was introduced in the first edition of Competition on Harvesting Raw Tables from Infographics (ICPR 2019 CHART Infographics)\cite{davila_19}. This dataset has synthetic images created using matplotlib. For the testing, a large set of synthetic data and a small set of real chart images harvested from PubMedCentral\footnote{https://www.ncbi.nlm.nih.gov/pmc/} were used. The training set has 198,010 images, whereas the synthetic test set has 4540 images, and the real test set has 4242 images. The dataset has ten different chart categories.

The second edition of the competition\cite{davila_20} provided a dataset containing 22923 real chart images of 15 different chart categories in both training and testing sets. The training set has 15636 images, while the test set has 7287 images. The annotations for the chart image samples are provided in both JSON and XML formats.
The dataset presented as a part of the third and most recent competition comprises 36183 images of 15 different chart categories. The training set contains 22,923  images, while the test set contains 13,260 images. Similar to the previous edition, the annotations are provided in JSON and XML formats.
To the best of our knowledge, this is the largest publicly available dataset for chart image classification.

The existing classification data sets for charts are summarized in Table \ref{tab:datasets}, and the composition of the publicly available datasets is reported in Table \ref{tab:Table3}.

\section{Deep Learning Models for Comparative Analysis}
In this section, we briefly discuss prominent deep-learning models that have been used to study the performance of chart classification. We have selected two categories of deep learning models - CNN-based and Transformer-based for the comparative study. For CNN-based models, we have considered the proven state-of-the-art models for image classification on the large-scale benchmark dataset ImageNet\cite{russakovsky_15} over the years. For vision transformer models, we have chosen the models that have been proven to outperform CNN-based models in computer vision tasks.

\subsection{ResNet\cite{he_15_resnet}}
The Deep Residual Network was introduced in 2015 and was significantly deeper than the previous deep learning networks. The motivation behind the model was to address the degradation problem: Degrading training accuracy with increasing depth of the model. The authors added shortcut connections, also known as skip connections, that perform the proposed identity mapping and are significantly easier to optimize than unreferenced mappings. Despite being deeper than the previous models, ResNet still needed to be simplified. It achieved the top-5 error of 3.57\% and claimed the top position in the 2015 ILSVRC classification competition\cite{russakovsky_15}. We use a 152-layer version of this Deep Residual Network called ResNet-152 for our classification problem.

\subsection{Xception\cite{chollet_17_xception}}
Xception is a re-interpretation of the inception module. The said inception module is replaced with depth-wise separable convolutions. The number of parameters in both Inception V3 and Xception is the same, so the slight performance improvement is due to the more efficient use of parameters. Xception shows a better performance improvement than Inception V3 on the JFT dataset on the ImageNet dataset. It achieves the top five accuracy of 94.5\%. Xception also shows promising results in the chart classification literature, as reported by \cite{araujo_20} and \cite{thiyam_22}.

\subsection{DenseNet\cite{huang_18_densenet}}
The Dense Convolutional Network, introduced in 2018, connects each layer in the network architecture to all other layers. This allows for the exchange of feature maps at every level and considers the same input as input gathered from all the previous layers rather than just one preceding layer. The difference between DenseNet and Resnet lies in the way that they combine features. ResNet combines features through summation, whereas DenseNet combines them through concatenation. DenseNet is easier to train due to the improved flow of gradients and other information through the network. The vanilla DenseNet has fewer parameters than the vanilla ResNet network. We used DenseNet-121 for our classification task as it was one of the best models for the chart image dataset as reported in \cite{thiyam_22}.

\subsection{ConvNeXt\cite{liu_22_convnet}}
ConvNeXt model was introduced as a response to hierarchical transformers outperforming convnets in image classification tasks. Starting with a standard ResNet architecture, the model is carefully modified to adapt the specific characteristics of a typical hierarchical transformer. This resulted in a CNN-based model that matches the transformers in robustness and scalability across all benchmarks. ConvNeXt achieves a top-1 accuracy of 87.8\% on ImageNet. 

\subsection{DeIT Transformer\cite{touvron_21_deit}}
The authors proposed the Data Efficient Image Transformer(DeIT) with 86M parameters to make the existing vision transformer more adoptable. This convolution-free approach achieves competitive results against the existing state-of-the-art models on ImageNet. The proposed vision transformer achieved a top-1 accuracy of 85.2\% on the ImageNet classification task. We use the base Base DeIT transformer for the chart classification task.

\subsection{Swin Transformer\cite{liu_21_swin}}
A hierarchical transformer that employs shifting windows to obtain representations for vision tasks. The authors note that the hierarchical architecture provides linear computational complexity and scalability concerning image size. The limitation of self-attention calculation concerning noncoincident local windows due to the shifting windows allows for better cross-window connectivity. The qualities above contribute to the Swin transformer's excellent performance across computer vision tasks. It achieves 87.3\% top-1 accuracy on the ImageNet-1k dataset. We perform experiments with all the 13 available Swin Transformer models and report their performance in Table \ref{tab:TableSwin}. Furthermore we refer to the best performing Swin Transformer model as Swin-Chart in Table \ref{tab:Table4}.

\begin{figure*}
    \centering
    \begin{tabular}{c c c c c} 
    \includegraphics[width=2 cm, height=2 cm]{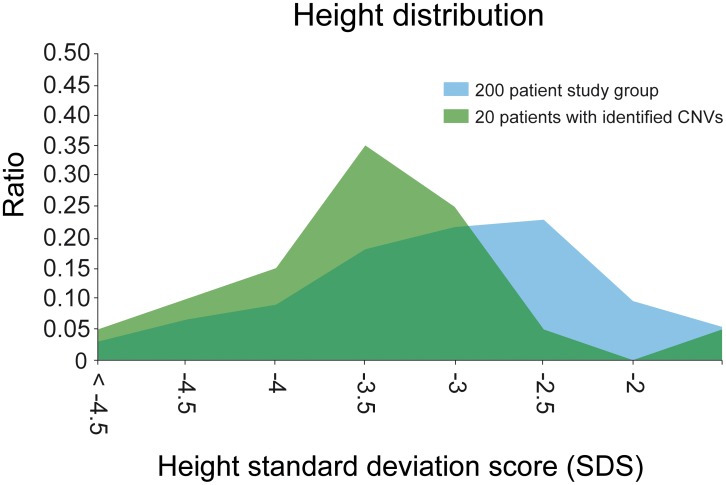} & 
    \includegraphics[width=2 cm, height= 2 cm]{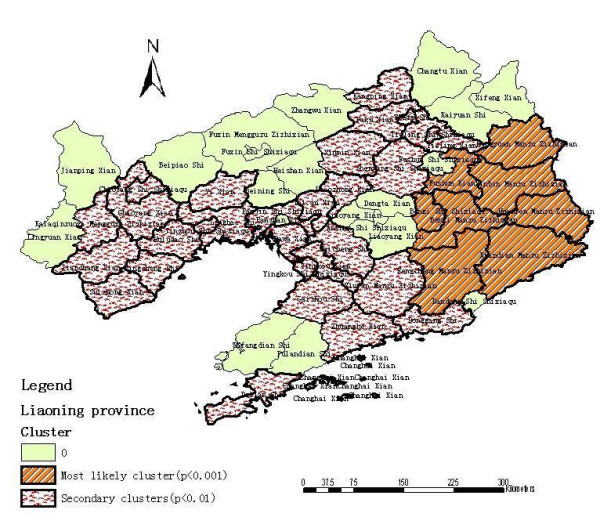} &
    \includegraphics[width=2 cm, height=2 cm]{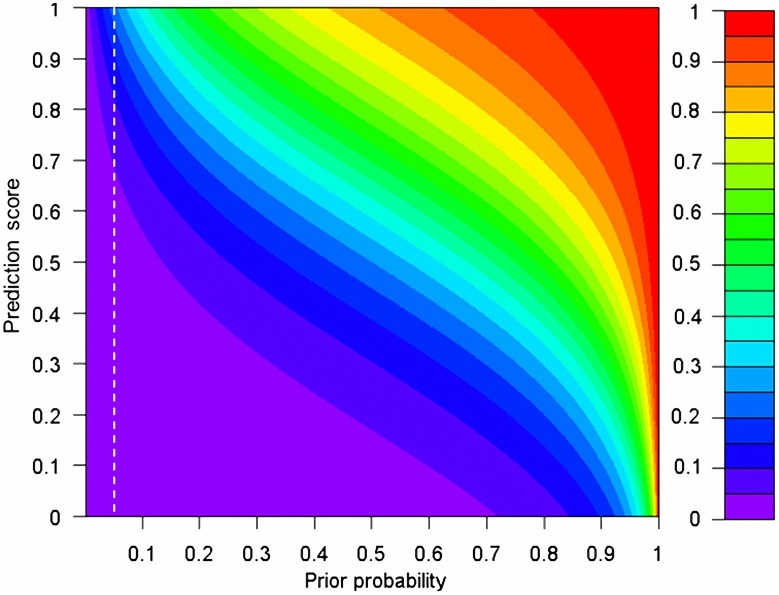} &
    \includegraphics[width=2 cm, height=2 cm]{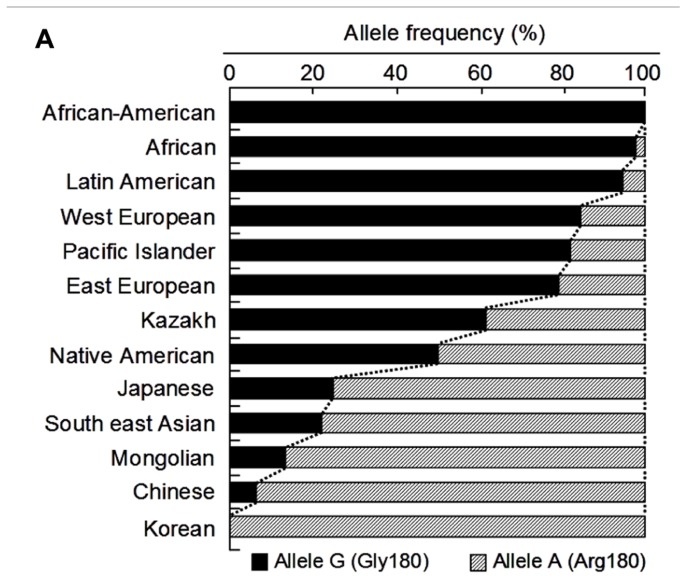} &
    \includegraphics[width=2 cm, height=2 cm]{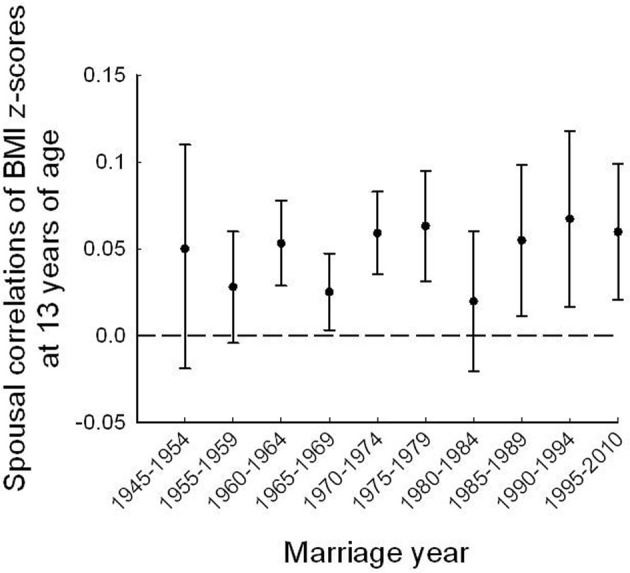} 
    \\
    \texttt{a. Area}  & \texttt{b. Map} & \texttt{c. Heatmap} & \texttt{d. Hor. Bar} & \texttt{e. Hor. Interval} 
    \\
    \\
    
    \includegraphics[width=2 cm, height=2 cm]{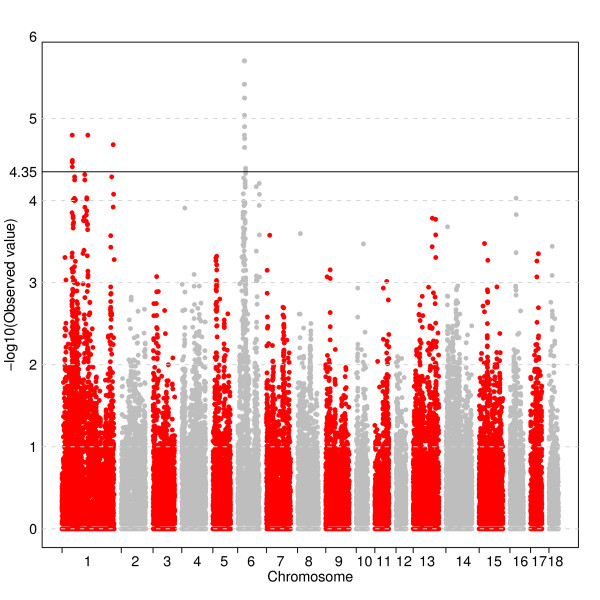} & 
    \includegraphics[width=2 cm, height= 2 cm]{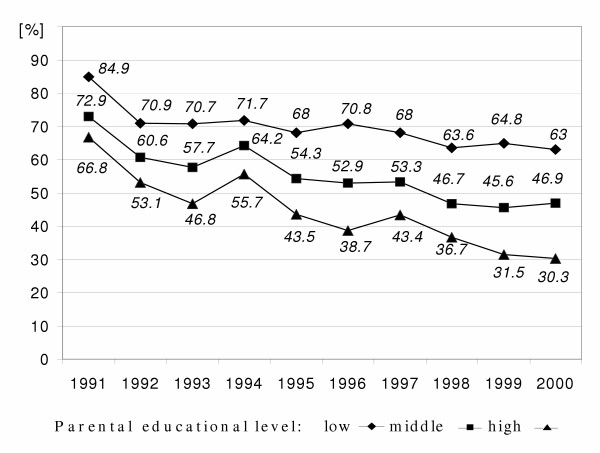} &
    \includegraphics[width=2 cm, height=2 cm]{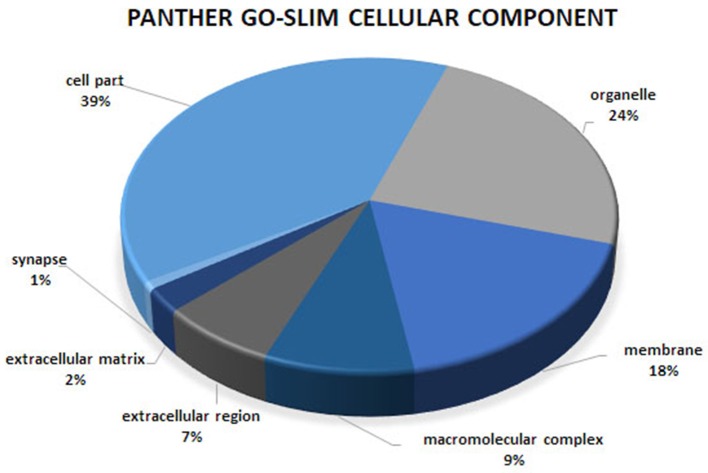} &
    \includegraphics[width=2 cm, height=2 cm]{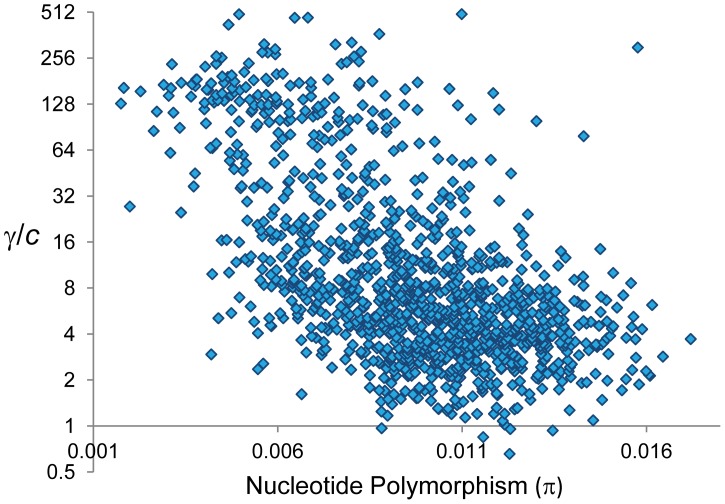} &
    \includegraphics[width=2 cm, height=2 cm]{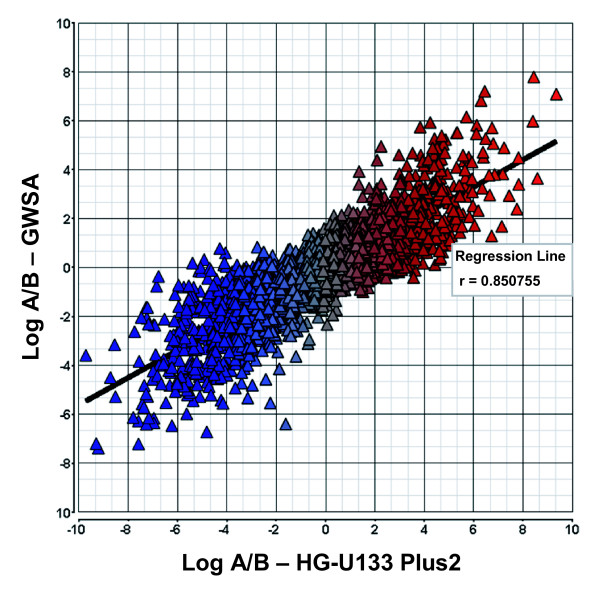} 
    \\   
    \texttt{f. Manhattan}  & \texttt{g. Line} & \texttt{h. Pie} & \texttt{i. Scatter} & \texttt{j. Scatter-line} 
    \\
    \\
    
     \includegraphics[width=2 cm, height=2 cm]{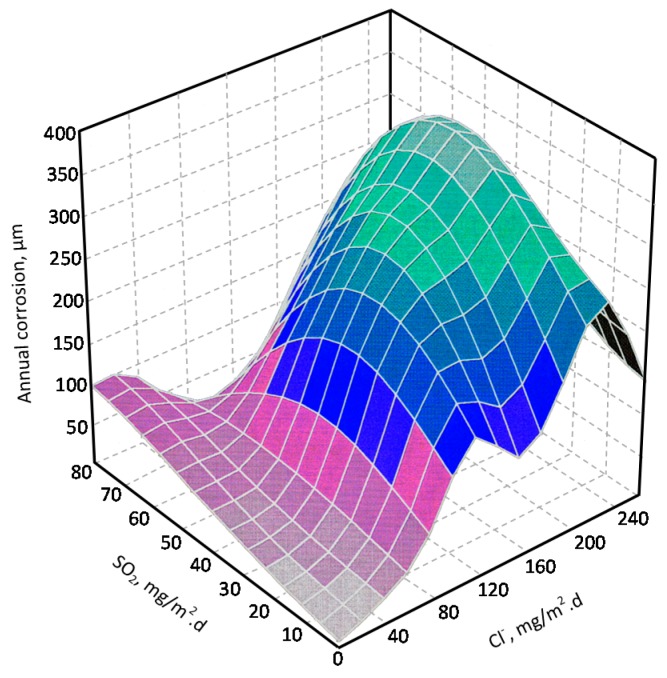} & 
    \includegraphics[width=2 cm, height= 2 cm]{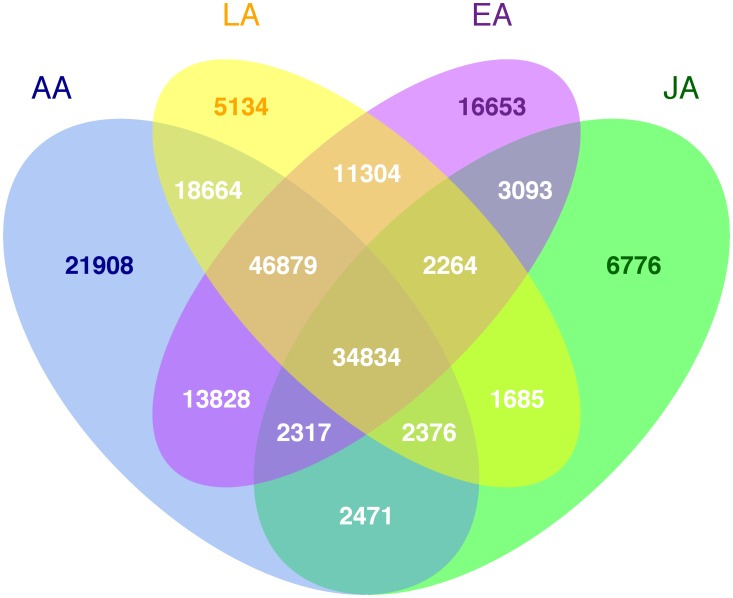} &
    \includegraphics[width=2 cm, height=2 cm]{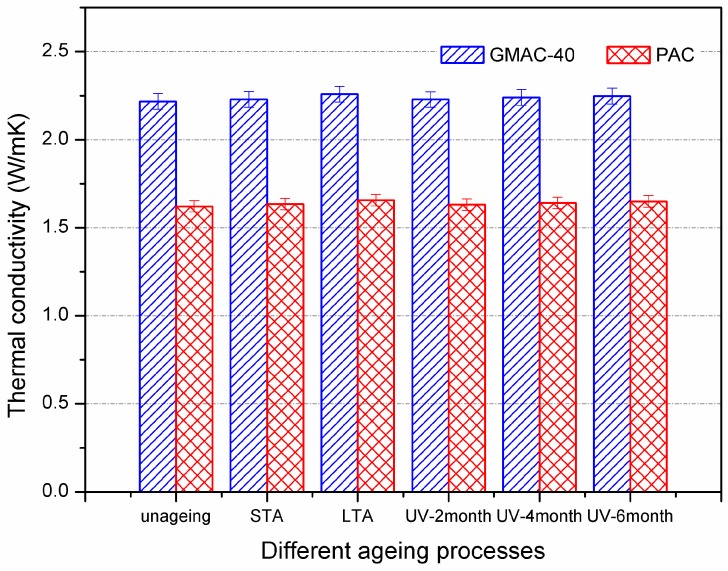} &
    \includegraphics[width=2 cm, height=2 cm]{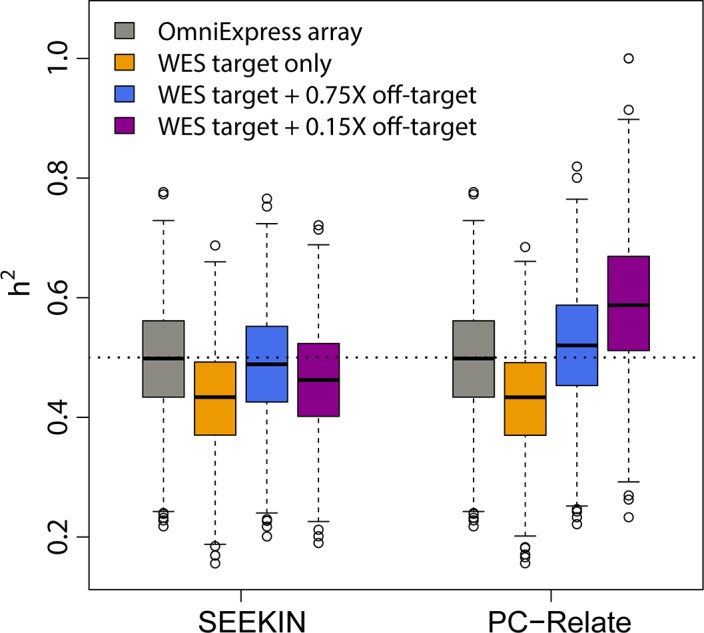} &
    \includegraphics[width=2 cm, height=2 cm]{Charts/PMC3458436___fgene-03-00125-g0002.jpg} 
    \\   
    \texttt{k. Surface}  & \texttt{l. Venn} & \texttt{m. Vert. Bar} & \texttt{n. Vert. Box} & \texttt{o. Vert. Interval} 
    \\   
    \\
    \end{tabular}
    \caption{Sample of chart images used in this study from UB-PMC\cite{davila_22} dataset}
    \label{fig:my_label}
\end{figure*}

\section{Experimental Protocol}
\subsection{Dataset}
We use the ICPR2022 CHARTINFO UB PMC\cite{davila_22} dataset to perform our comparative study of deep learning models. The dataset is divided into training and testing sets. The number of chart images in the training and test set is 22,923 and 11,388, respectively. The ground truth values are annotated in JSON and XML formats. We further divide the provided training set into training and validation sets with an 80/20 ratio. The dataset contains charts of 15 categories: area, map, heatmap, horizontal bar, Manhattan, horizontal interval, line, pie, scatter, scatter-line, surface, Venn, vertical bar, vertical box, and vertical interval. Samples of each chart type present in the dataset are shown in Figure \ref{fig:my_label}

\subsection{Training and Testing Setup}
We choose ResNet152, DenseNet121, Xception, and ConvNeXt CNN-based models and DeIT and Swin Transformers-based models for chart image classification. The CNN-based models were selected based on their performance in the existing literature on the ImageNet image classification task. The transformer-based models are chosen because they beat the CNN-based models. We use the pre-trained ImageNet weights of these models and fine-tune them for our chart classification task. The models are trained on a computer with an RTX 3090 video card with 24 GB memory. Pytorch\cite{paszke_19} was used as the engine for our experiments. We use a batch size of 64 for CNN-based models and a batch size of 16 for transformer-based models. A learning rate of $10^{-4}$ is used to train each model for 100 epochs. Label Smoothing Cross Entropy Loss is used as a loss function. The evaluation measures the average over all classes and reports precision, recall, and F1-score.

        \begin{table}[h]
            \centering
            \caption{Comparative Performance of all the 13 Pre-trained Swin Transformer Models on ICPR2022 CHARTINFO UB PMC datase}
            \begin{tabular}{|l|c|c|c|}
            \hline
                 Model & Precision & Recall & F1-measure \\
                 \hline
                 \hline
                 SwinT & 0.929 & 0.924 & 0.922\\
                 SwinT\_s3 & 0.931 & 0.923 & 0.922\\
                 SwinS & 0.931 & 0.926 & 0.925\\
                 SwinS\_s3 & 0.928 & 0.922 & 0.919\\
                 SwinB\_224 & 0.933 & 0.926 & 0.925\\
                 SwinB\_384 & 0.936 & 0.932 &  0.931\\
                 SwinB\_224\_in22k\_ft1k & 0.934 & 0.930 & 0.929\\
                 SwinB\_384\_in22k\_ft1k & 0.933 & 0.929 & 0.927\\
                 SwinB\_s3 & 0.927 & 0.923 & 0.921\\
                 \textbf{SwinL\_224} & \textbf{0.937} & \textbf{0.933} & \textbf{0.932}\\
                 SwinL\_384 & 0.937 & 0.931 & 0.929\\
                 SwinL\_224\_in22k\_ft1k & 0.937 & 0.933 & 0.932\\
                 SwinL\_384\_in22k\_ft1k & 0.934 & 0.930 & 0.929\\
                 \hline
            \end{tabular}
            \label{tab:TableSwin}
        \end{table}

        \begin{table}[h]
            \centering
            \caption{Comparative performances of the CNN-based and Transformer-based models on ICPR2022 CHARTINFO UB PMC dataset}
            \label{tab:Table4}            
            \begin{tabular}{|l|c|c|c|}
            \hline
                 Model & Precision & Recall & F1-score\\
            \hline
            \hline
                 Resnet-152\cite{he_15_resnet} & 0.905 & 0.899 & 0.897\\
                 Xception\cite{chollet_17_xception} & 0.882 & 0.870 & 0.866\\
                 DenseNet-121\cite{huang_18_densenet} & 0.887 & 0.879 & 0.875 \\
                 ConvNeXt\cite{liu_22_convnet} & 0.906 & 0.898 & 0.896\\
                 DeIT\cite{touvron_21_deit} & 0.888 & 0.879 & 0.874\\
                 Swin-Chart & \textbf{0.937} & \textbf{0.933} & \textbf{0.932} \\
            \hline
            \end{tabular}
        \end{table}

        \begin{table}[h]
            \centering
            \caption{Comparison of Swin-Chart from Table \ref{tab:Table4} with models stated in \cite{davila_22} on the ICPR2022 CHARTINFO UB PMC dataset}
            \label{tab:Table5}
            \begin{tabular}{|l|c|c|c|}
            \hline
                 Team & Precision & Recall & F1-score\\
            \hline
            \hline
                 Our (Swin-Chart) & \textbf{0.937} & \textbf{0.933} & \textbf{0.932} \\
                 IIIT\_CVIT & 0.926 & 0.901 & 0.910\\
                 UB-ChartAnalysis & 0.900 & 0.881 & 0.886\\
                 six seven four & 0.865 & 0.808 & 0.827\\
                 CLST-IITG & 0.704 & 0.657 & 0.654\\

            \hline
            \end{tabular}
        \end{table}

\subsection{Comparative Results}
The models were trained following the steps mentioned in the previous section and were tested on the UB-PMC test data set. We calculate all deep learning models' average precision, recall, and F1 score. Among CNN-based models, ResNet-152 and ConvNeXt provide the best results across all evaluation metrics. The ResNet-152 result is consistent with the results in \cite{davila_22} for CNN-based models. For Swin transformer we perform experiments on 13 models consisting Swin Tiny(SwinT), Swin Small(SwinS), Swin Base(SwinB) and Swin Larger(SwinL) and their varients. SwinL with input image dimension 224 performs best with an F1-score of 0.932. So, \textbf{SwinL} model is further referred as \textbf{Swin-Chart}. The scores of all the Swin Transformer models are summarized in Table \ref{tab:TableSwin}. The best performing CNN based models fail to compete with Swin-Chart for the chart classification task as it outperforms the other five models with an average F1-score of 0.932. The scores for the deep learning models are summarized in Table \ref{tab:Table4}. 

Furthermore, we compare our best-performing model(Swin-Chart) with the models reported in \cite{davila_22}. This comparison is summarized in Table \ref{tab:Table5}. We note that Swin-Chart surpasses the winner of the ICPR 2022 CHART-Infographics competition with an average F1-score of 0.931.

\section{Future Directions}
Although there has been a significant increase in published articles on chart classification, several problems still need to be addressed. 
\subsection{\textbf{Lack of Standard Benchmark Data Sets}}
The chart image classification problem has been extensively addressed in previous work. Efforts have been made to increase the size of chart image datasets that also cover a wide variety of charts\cite{davila_19,thiyam_22}. With the growing literature in various domains, authors are finding creative ways to use different charts. This adds to the variety of chart types. Integrating such diverse chart types while creating chart datasets remains an open challenge. In addition, the popularity of charts such as bar, line, and scatter over others such as Venn, surface, and area adds to the problem of disparity between the number of samples in particular chart types.  

\subsection{\textbf{Lack of Robust Models}}
Recent work makes some problematic assumptions in addressing this problem\cite{davila_19_survey}. A lack of a diverse benchmark dataset adds to this problem, as there needs to be more consistency in model performance across publicly available datasets. The inherent intra-class dissimilarity and inter-class similarity of several chart types affect the model's performance.

\subsection{\textbf{Inclusion of Noise}}
Most of the work in the existing literature ignores the effect of noise. Different types of noise, such as background grids, low image quality, composite charts, and multiple components along with figures, lead to poor performance for models that perform exceptionally well on noiseless data\cite{thiyam_21}. In addition to the noiseless chart image dataset, if a small set of chart images could be provided that incorporates the noisy images, it would help fine-tune the models to work through the inclusion of noise and be invariant to the same.   

\section{Conclusion}
We have provided a brief survey of existing chart classification techniques and datasets. We used a Transformer model to obtain state-of-the-art results. Although there has been a significant development both in terms of variety in models and in the size of datasets, we observe that the chart classification problem still needs to be solved, especially for noisy and low-quality charts. Our comparative study showed that Swin-Chart outperforms the other vision transformer and CNN-based models on the latest UB-PMC dataset. In the future, we plan to generalize the results of the Swin-Chart over other publicly available datasets and try to bridge the gap to a robust deep-learning model for chart image classification.

%
%
%
%

\end{document}